\pdfoutput=1

\documentclass[11pt]{article}

\usepackage[]{ACL2023}

\usepackage{comment}
\usepackage{times}
\usepackage{latexsym}
\usepackage{graphicx} 
\usepackage[T1]{fontenc}

\usepackage[utf8]{inputenc}

\usepackage{microtype}

\usepackage{amsmath}
\usepackage{inconsolata}
\usepackage{algpseudocode}
\usepackage{algorithm}
\usepackage{multirow}
\usepackage{multicol}
\usepackage{amsfonts}
\usepackage{makecell}
\usepackage{multirow}

\usepackage{enumitem}

\title{LEGO: Language Model Building Blocks}

\author{ Shrenik Bhansali, Alwin Jin, Tyler Lizzo, Larry Heck\\ AI Virtual Assistant (AVA) Lab\\
   Georgia Institute of Technology  \\ \texttt{\{sbhansali8,ajin37,lizzo,larryheck\}@gatech.edu}}

\begin{document}
\maketitle
\begin{abstract}
Large language models (LLMs) are essential in natural language processing (NLP) but are costly in data collection, pre-training, fine-tuning, and inference. Task-specific small language models (SLMs) offer a cheaper alternative but lack robustness and generalization. This paper proposes LEGO, a novel technique to extract SLMs from an LLM and recombine them. Using state-of-the-art LLM pruning strategies, we can create task- and user-specific SLM building blocks that are efficient for fine-tuning and inference while also preserving user data privacy. LEGO utilizes Federated Learning and a novel aggregation scheme for the LLM reconstruction, maintaining robustness without high costs and preserving user data privacy. We experimentally demonstrate the versatility of LEGO, showing its ability to enable model heterogeneity and mitigate the effects of data heterogeneity while maintaining LLM robustness. \textbf{Our codebase will be released upon publication.}

\end{abstract}

\section{Introduction}

Large Language Models (LLMs) represent a significant advance  in 
Natural Language Processing (NLP) with their remarkable ability to generalize across queries and tasks. These models are typically fine-tuned using large, diverse datasets derived from high-quality instruction data \cite{gupta2022recovering}. 

LLMs are not, however, a one-size-fits-all solution. Running LLMs on small devices like IoT devices or smartphones is not possible due to their resource limitations. Downstream LLM applications that prioritize privacy, such as personal conversational AI, become untenable due to data privacy concerns, as user data must stay on personal devices or private networks and cannot be shared globally. These privacy constraints apply to both fine-tuning and inference.

LLMs are traditionally fine-tuned in a centralized manner, where data is aggregated from raw user interactions and shared globally to fine-tune a single global model. In contrast, 
Federated Learning (FL) is a collaborative learning approach that allows client models to learn from users while preserving their privacy \cite{mcmahan2017communication}.  
FL utilizes distributed fine-tuning with localized client models trained on localized user interactions, resulting in a global model created by aggregating client model weights. While FL preserves data privacy and addresses the complexity of fine-tuning, it does not resolve the high cost of inference with LLMs.

Small Language Models (SLMs) address the high cost of inference and fine-tuning, allowing for a greater range of client devices. While SLMs are more efficient, the cheaper performance comes at the expense of robustness and generalization across broad tasks, conversational interactions, and advanced LLM capabilities. Furthermore, SLMs are not typically designed to be composable, constraining FL architecture to an either-or choice: choose SLMs at the cost of robustness, or choose the original LLMs that limit their utility due to size and complexity.

For resource-constrained scenarios like chatbots on small devices, there is a critical need for computationally efficient, robust, general, and private methods that facilitate different sizes and architectures of models depending on the computational resources of the device.

To enhance client flexibility in distributed conversational AI systems, we introduce {\bf L}anguage Mod{\bf E}l Buildin{\bf G} Bl{\bf O}cks (LEGO), a model-agnostic technique for federating small language models (SLMs) with diverse heterogeneous architectures. LEGO enables efficient fine-tuning and inference, preserves privacy, optimizes performance across varied resource constraints, and aids in developing robust and generalizable large language models (LLMs). Our approach utilizes an SLM-based FL system where each SLM is derived from an LLM, allowing them to be combined to reconstruct the original LLM. By treating SLMs as building blocks, LEGO effectively assembles them into a cohesive LLM.

Through the use of LEGO, we demonstrate a flexible FL system that broadens the range of possible client devices by enabling different-sized models for different-sized devices. Experiments show that when using LEGO, SLMs are better learners and therefore yield more robust models. We also demonstrate that SLMs can better adapt to data heterogeneity when compared to LLMs. Through LEGO, we can leverage the advantages of SLMs, and treat them as composable building blocks that combine to form an LLM.

\vspace*{.1in}
\noindent With the proposed LEGO approach, the major contributions of this work include 
\vspace*{-.1in}
\begin{itemize}[itemsep=1mm, parsep=0pt]
    \item A method to compose SLMs together to yield a robust and generalizable LLM
    \item A privacy-preserving FL architecture to serve these composable client-side heterogeneous SLMs
    \item A method to optimize client-side SLMs against heterogeneous resource budgets for efficient fine-tuning and inference
\end{itemize}

The rest of this paper is organized as the following: 
Section~\ref{background} gives background information. 
Section~\ref{method} details the methodology behind the LEGO approach and its components.
Section~\ref{experiments} covers the experiments we performed to validate LEGO and houses their results.
Section~\ref{related} discusses the related work. 
Section~\ref{conclusions} concludes the paper and Section~\ref{limits} lists our study's limitations.

\section{Background}\label{background}

\subsection{Model Compression}\label{compression}

In recent years, pruning has become widely used in NLP to compress LLMs~\cite{lecun1989optimal}. Pruning involves the selective omission of model parameters with minimal contributions to the learning process. Pruning techniques have proven successful, enhancing the cost-effectiveness of large pre-trained models \cite{xia2023sheared}.

Recently, more nuanced pruning approaches have been discussed in the literature, improving over more traditional methods like magnitude pruning. Specifically, two state-of-the-art pruning methods are widely discussed in the literature---SparseGPT~\cite{frantar2023sparsegpt} and Wanda~\cite{sun2023simple}. Whereas traditional magnitude pruning operates by pruning weights with the largest magnitude, these pruning techniques instead track weight activations, and prune weights with the lowest amount of activation.

SparseGPT creates and solves a layer-wise reconstruction problem to determine the weight activations, whereas Wanda takes the product of a weight's magnitude and the norm of its associated input activations.


\begin{figure*}[t]
    \centering
    \includegraphics[width=.7\linewidth]{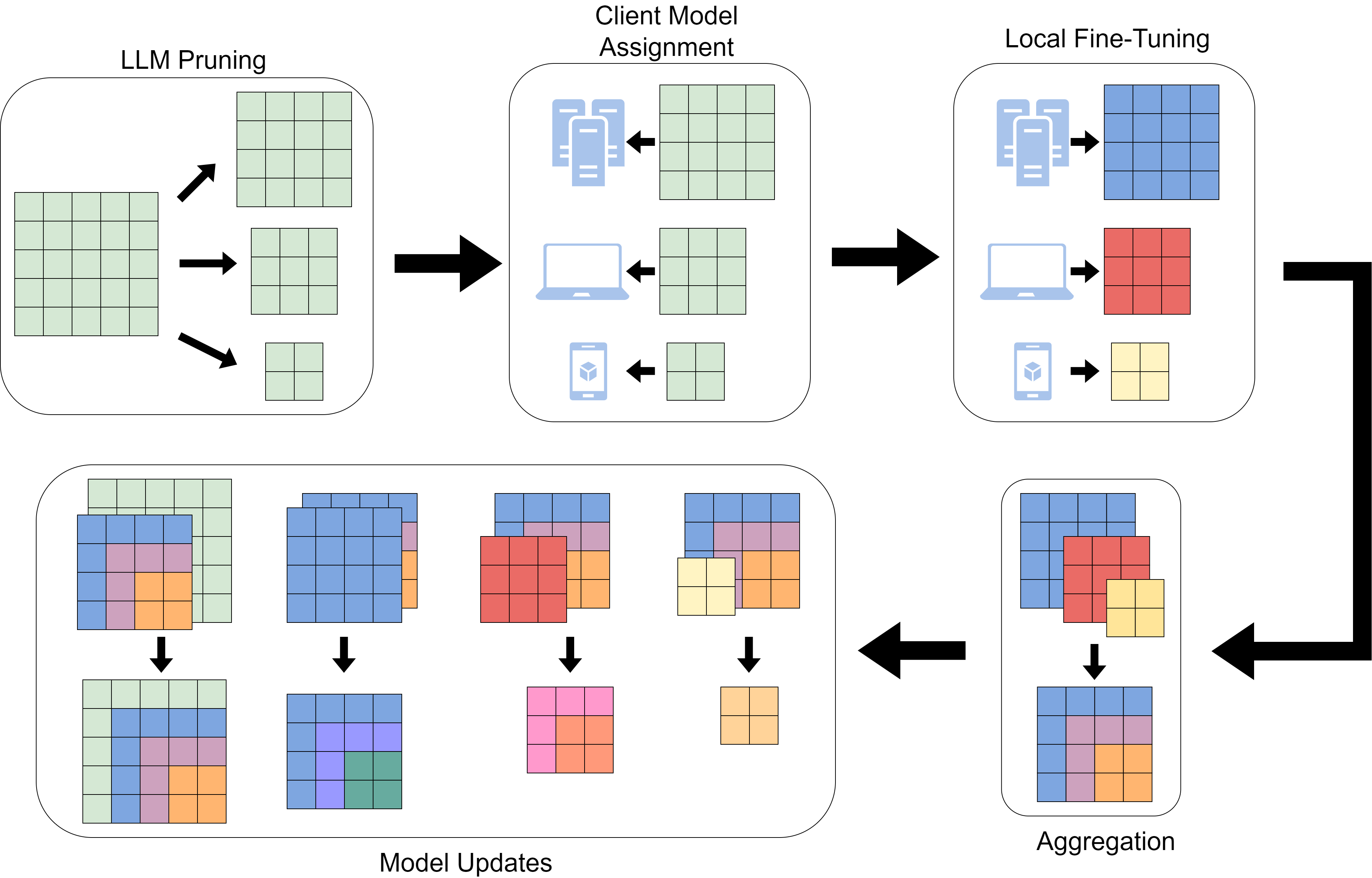}
    \caption{The LEGO workflow. An LLM is first pruned to create SLMs, then each SLM is assigned to a client. Each client then fine-tunes its SLM on its local data. After fine-tuning, the models are aggregated to create a global update. The global update is then applied to all the client SLMs as well as a global LLM. Eventually, after enough updates, a final global LLM is derived. 
    }
    \label{fig:LEGO}
\end{figure*}

\subsection{Federated Fine-Tuning}

Federated Learning (FL) is a distributed training methodology that trains a model across multiple decentralized devices while allowing data to remain on user machines~\cite{mcmahan2017communication}. In conventional FL, each client device has its own native model and trains it on local user inputs. Instead of sharing this client data globally, the models instead share their own model weights, aggregating them with other client weights. This creates a global update that encodes knowledge gained from all model updates without compromising data privacy.

FedIT is a technique \cite{zhang2023towards} which leverages FL as the learning framework for the instruction tuning of LLMs. FedIT is motivated by the recent success of instruction-tuned generative large language models on generalizing to new tasks. This FL method mitigates the dependence and associated cost, accessibility, and lost privacy of instruction-tuned LLMs on large amounts of centralized high-quality human-written instructions by federating the learning phases.

Two fundamental assumptions are often made in both traditional FL and the fine-tuning of LLMs with FL (FedIT). The first is that all data is i.i.d., meaning that not only do all clients have the same amounts of data, but that the ratio of content within each is the same. The study of non-i.i.d. data distributions in FL is often referred to as heterogeneous FL~\cite{ye2023heterogeneous, ghosh2019robust}. 

The second assumption is that all model architectures in FL systems are identical, allowing for the aggregation of model weights when creating global updates.
Heterogeneity in model architecture therefore presents unique challenges in FL, such as impeding the use of aggregation techniques like FedAvg \cite{li2019convergence} that average the federated weights assuming homogeneity. 

We seek an approach that not only leverages the distributed nature of FL to preserve user data privacy on client devices but also allows for each client to host heterogenous SLMs and heterogeneous data sets of various sizes and makeup (e.g., task-dependent). This new architecture would facilitate the optimization of client-side SLMs against heterogeneous resource budgets in both data size and compute. 

\section{Methodology}\label{method}

Motivated by the need for efficient fine-tuning and inference for private, resource-constrained scenarios, we propose a model-agnostic FL system {\bf L}anguage  Mod{\bf E}l Buildin{\bf G} Bl{\bf O}cks (LEGO). Much like stacking small building blocks together to create a larger structure, we stack small language models (SLMs) together to create a larger, more robust Large Language Model (LLM). 

LEGO employs a two-step approach. First, we obtain SLMs of different sizes by pruning an LLM. We then deploy these SLMs in an FL environment, eventually aggregating them into an LLM. $\mbox{Figure \ref{fig:LEGO}}$ shows the LEGO workflow in greater detail. The SLMs produced by the pruning process are the local client models in the FL environment. We produce SLMs of different sizes to better match the various computational budgets of client devices. We use a full-sized LLM as the global model, meaning that every client model is a sub-network of the global model.

To produce a fine-tuned LLM using the client SLMs, we begin the process of federated fine-tuning. First, the selected client SLMs for each round are fine-tuned on their respective client's local data. They are then aggregated with each other, creating a global update. This global update is then applied to all client SLMs and the global LLM. We repeat this process for every round of FL, eventually forming a robust, fine-tuned LLM built from the updates supplied by the fine-tuned client SLMs. 

For all studies and experiments, we impose the following conditions:
\vspace*{-.1in}
\begin{itemize}[itemsep=1mm, parsep=0pt]
    \item All fine-tuning is done using LoRA \cite{hu2021lora}, resulting in a more computationally efficient fine-tuning process. The LoRA adapters preserve model sparsity. We provide more configuration details in Appendix \ref{expsetup}.
    \item All aggregation occurs over the LoRA adapters, allowing for decreased communication cost and more efficient aggregation.
    \item All fine-tuning is done over the databricks-dolly-15k dataset or a subset of it. This dataset was generated by Databricks and covers eight different capability domains from the InstructGPT paper~\cite{ouyang2022training}.
\end{itemize}

\subsection{Model Pruning}
For our experiments, we simulate an FL system on our cluster. We examine 4 model sparsity levels (0\%, 25\%, 50\%, and 75\%), where each percentage indicates the proportion of weights that have been removed. To create the SLMs, we use SparseGPT~\cite{frantar2023sparsegpt} to remove the weights from a LLaMA-7B LLM, inducing the specified level of sparsity in each model.
We compare SparseGPT against Wanda and Layer pruning in \ref{subsec:appenpruning} and determine that SparseGPT is the strongest pruning strategy for LEGO.

\subsection{Model-agnostic Federated Learning}

If SLMs are the building blocks, then FL is the process of assembling the blocks into a structure, and the resulting global LLM is the final, completed structure. We create a model-agnostic FL environment to allow aggregation between different sized SLMs, and the global LLM. At the end of the FL process, we obtain a fine-tuned global LLM constructed through the aggregation of SLMs.

\begin{algorithm}[h!]
    \caption{Federated Fine-Tuning with \\Heterogeneous Models}
    \label{alg:fed_tuning_sparsity}
    \begin{algorithmic}
    \small
    \State \textbf{Initialization}:
    \State Each client $n$ initializes LLM with parameter sparsity $w_n$.
    \State $M \gets \emptyset$; $K$ communication rounds; $k \gets 0$.
    
    \State \textbf{Training Loop}:
    \While{$k \leq K$}
        \State Update $M$ to select clients based on sparsity.
        \For{each client $n \in M$}
            \State Select model for $n$ with $w_n$.
            \State $\Delta w_{k+1,n} \gets \text{InstructionTune}(\Delta w_{k,n})$.
        \EndFor
        \State $\Delta w_{k+1} \gets \text{HeteAgg}(\{\Delta w_{k+1,n} : n \in M\})$.
        \State $k \gets k + 1$.
    \EndWhile

    \State \textbf{Outcome}:
    \State Derive final adapters $\Delta w_K$; update global LLM $w$.
    \end{algorithmic}
\end{algorithm}

Algorithm \ref{alg:fed_tuning_sparsity} details our FL system, where clients would be assigned their respective SLMs with $w_n$ sparsity, representing the sparsity present in both the model and the LoRA adapter. 
During the training loop, clients fine-tune their LoRA adapters on local data created from a subset of the databricks-dolly-15k dataset. After fine-tuning, each of the selected clients has their LoRA adapters aggregated with each other to form a global update through the HeteAgg method---our heterogeneous model aggregation scheme detailed in Algorithm \ref{alg:HeteAgg} . This global update is then applied to each of the client SLMs in addition to the global LLM. After the training loop is complete, we can derive our final adapters and global updates.

\begin{algorithm}[h!]
\caption{Model Heterogeneous Aggregation (HeteAgg)}
\label{alg:HeteAgg}
\begin{algorithmic}
\small
\State Load initial global model state dictionary: $g$
\State Define the number of clients $n$.
\State Derive global parameter sums $\mathcal{P}_{\text{sums}}$ \& counts $\mathcal{P}_{\text{counts}}$.
\For{each client $i \in \{1, \dots, n\}$}
    \State Load client model state dictionary: $s_i$
    \State Identify $\mathcal{P}_{g}$, the set of global model parameters.
    \For{each parameter $p \in \mathcal{P}_{g}$}
        \State Load $p_i$ from $s_i$
        \State Define mask $M_i \leftarrow (p_i \neq 0)$
        \State Update $\mathcal{P}_{\text{sums}}[p] \leftarrow \mathcal{P}_{\text{sums}}[p] + \text{where}(M_i, p_i, 0)$
        \State Update $\mathcal{P}_{\text{counts}}[p] \leftarrow \mathcal{P}_{\text{counts}}[p] + M_i$
    \EndFor
\EndFor
\For{each parameter $p \in \mathcal{P}_{g}$}
    \State $p_{\text{avg}} \leftarrow \mathcal{P}_{\text{sums}}[p] / \max(\mathcal{P}_{\text{counts}}[p], 1)$
    \State Update global model with $p_{\text{avg}}$
\EndFor
\For{each client $i \in \{1, \dots, n\}$}
    \For{each parameter $p \in \mathcal{P}_{g}$}
        \State Load $p_i$ from $s_i$
        \State Define mask $M_i \leftarrow (p_i \neq 0)$
        \State $s_i[p] \leftarrow \text{where}(M_i, p_{\text{avg}}, p_i)$
    \EndFor
\EndFor
\end{algorithmic}
\end{algorithm}

In our HeteAgg approach, we begin by instantiating a global LLM to hold the eventual global update. This global update is formed by aggregating the client SLMs. This is done by accessing each of the selected client's LoRA adapters, and creating a mask for it based on its sparsity. This sparse mask is then aggregated with the global LLM's LoRA adapter wherever there is overlap between the mask and the adapter. Since sparsity is represented by a parameter magnitude '0' in the SLM's LoRA adapters, this process effectively averages the nonzero parameters between the client and global models. 

By only aggregating across the nonzero weights, we can retain the sparsity in the client model's adapter without halving the global adapter's weights when there is no corresponding nonzero value. This process of mask creation and aggregation occurs for every client in the selected client group, forming a global update through the global LLM's adapter. Since every client SLM is a sub-model of the LLM, we can apply the global update to each client in the same manner again using HeteAgg, averaging across each client's nonzero weights.

\begin{figure}[h]
    \centering
    \includegraphics[width=\linewidth]{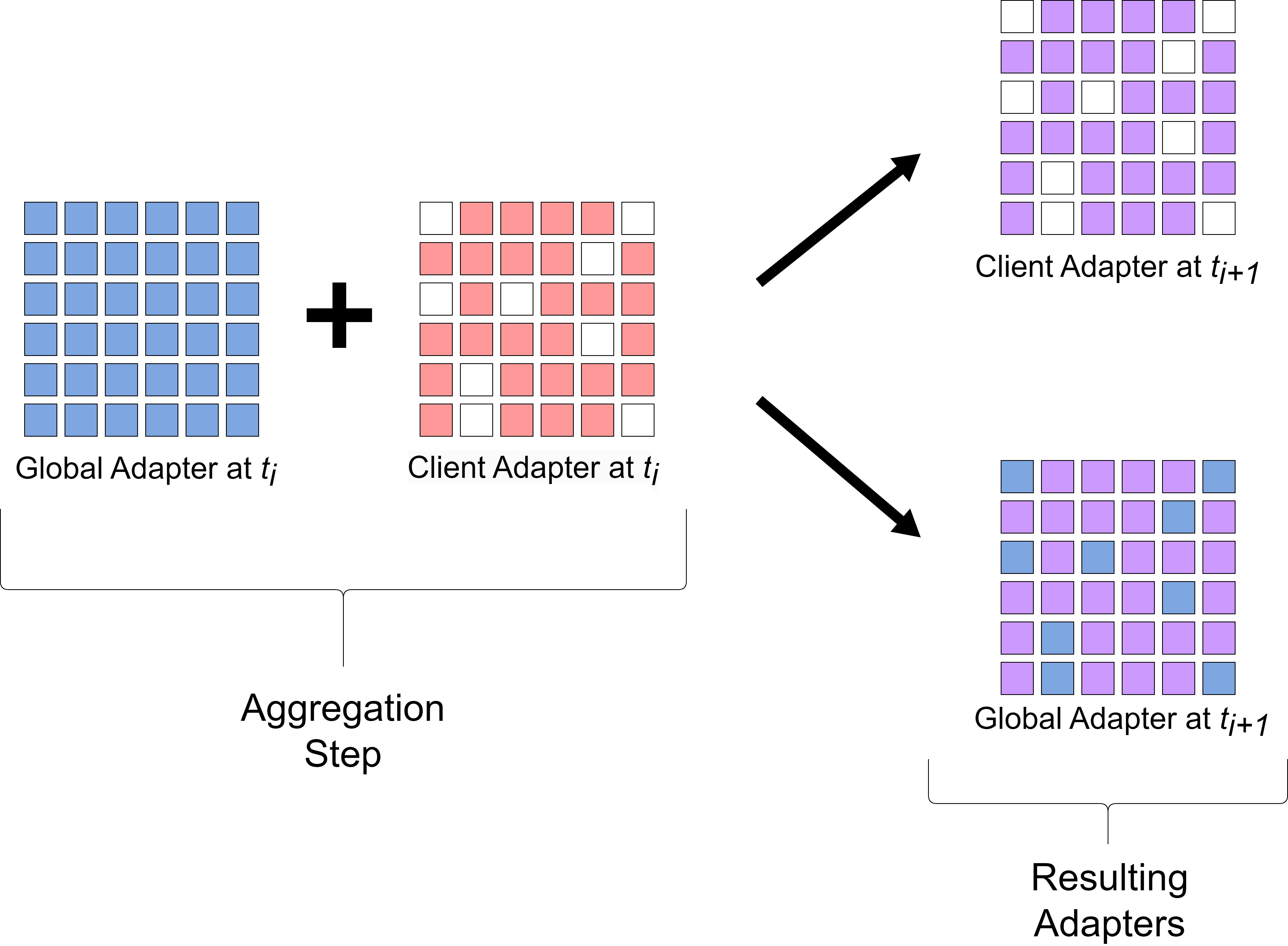}
    \caption{A symbolic representation of our heterogeneous aggregation method. 
    }
    \label{fig:heteaggfig}
\end{figure}

Figure \ref{fig:heteaggfig} represents our heterogeneous aggregation method, where the blue matrix represents the global LoRA adapter, and the red matrix represents a sparsified client LoRA adapter. The left-hand side displays each adapter at timestep $t_i$, before aggregation. During aggregation, the blue and red parameters average to create purple parameters for non-zero red (client) parameters. For zero-valued red (client) parameters, the updated client model retains its sparsity (upper right matrix), whereas the updated global LoRA adapter uses the blue (global) parameter values. As a result, the updated global adapter is a 0\% sparsity adapter. Thus, the right-hand side displays each adapter at timestep $t_{i+1}$, where the parameters are aggregated only when there is an overlap between the corresponding non-zero parameters of each model.

\section{Experiments}\label{experiments}

To examine the efficacy of our LEGO methodology, we conduct experiments to answer the following questions:
\begin{itemize}
    \item With i.i.d. {\bf task-independent data} and client-distributed SLMs as LEGO blocks, does the recombination of the SLMs yield a robust LLM?
    \item What is the effect of fine-tuning each client SLM in LEGO  with non-i.i.d.  {\bf task-dependent data?} 
    \item Does LEGO enable the combination (stacking) of differently shaped blocks? 
\end{itemize}

In each experiment, we follow the LEGO workflow as illustrated in Figure \ref{fig:LEGO}. We first prune an LLM to create SLMs, instantiate the SLMs as client models, fine-tune each client on their respective local client data, then aggregate them together to form a global update. The global update is then applied to a 0\% sparsity global model and the client models, which are then evaluated.

We compare LEGO with these baselines:
\begin{itemize}
    \item A FedIT-produced global model resulting from  4 LLaMA-7B client models fine-tuned over i.i.d. data. This baseline is the ideal case for FedIT.
    \item A FedIT-produced global model resulting from 8 task-specific LLaMA-7B client models where each model is only fine-tuned on one of the 8 different domain areas of databricks-dolly-15k. 
\end{itemize}

For all LEGO experiments, we use our HeteAgg method to aggregate the client weights, accounting for their heterogeneity (as opposed to the simple averaging of all weights in FedAvg). Since the computational cost of HeteAgg is the same as FedAvg, all speedups in LEGO are a direct result of model pruning \cite{sun2023simple, frantar2023sparsegpt}. During our experiments, we observe up to a $1.6\times$ speedup in inference and up to a $1.4\times$ speedup in fine-tuning using SparseGPT-produced SLMs when compared to 0\% sparsity LLMs. 
Given the approach is decoupled from the specific pruning method, LEGO will see further speedups as pruning methods improve.

\subsection{LEGO with Task-Independent Data} \label{iidexp}
\setlength{\tabcolsep}{4pt}
\begin{table*}[t]
\centering
\caption{Average Model Accuracy Over Benchmarks}
\label{table:results}
\begin{tabular}{ccccc}
\hline
\textbf{Composition} & \makecell{\textbf{LLM}\\\textbf{Initial}} & \makecell{\textbf{SLMs}\\\textbf{Averaged Initial}} & \makecell{\textbf{SLMs}\\\textbf{Globally Updated}} & \makecell{\textbf{LLM}\\\textbf{Recombined}} \\ \hline
{\bf LEGO} 4 SLMs With i.i.d. Data  & 0.559 & 0.240 & 0.416 & 0.568 \\
\textbf{FedIT}: 4 LLMs With i.i.d. Data  & 0.559 & N/A & N/A & 0.567 \\ \hline
{\bf LEGO}  8 Task-Dependent SLMs & 0.559 & 0.240 & 0.411 & \textbf{0.571} \\
\textbf{FedIT}: 8 Task-Dependent LLMs & 0.559 & N/A & N/A & 0.563 \\ \hline
\end{tabular}
\end{table*}
When building large structures, it is common to assemble smaller sub-units individually before combining them into the final form. Similarly, with LEGO, we can fine-tune smaller models individually, treating them as sub-units that are then aggregated together to produce a final LLM.
This experiment tests the transferability of knowledge from SLMs to an untouched LLM using LEGO. 

We prune an LLM to create four 75\% sparsity SLMs with an i.i.d. client data distribution. 
The dataset is partitioned into 4 i.i.d. segments, such that each client dataset covers the same domains in the same ratios. Each round, one client will fine-tune off their local data, and then be aggregated with the global 0\% sparsity LLM. A global update is then derived and applied to all of the client models. 

The results of this experiment with 4 SLMs and i.i.d. data for fine-tuning are shown in Table \ref{table:results}. 
As expected, due to the i.i.d. nature of the data, LEGO matches FedIT's accuracy on the recombined 0\% sparsity global LLM (shown in the last column, 'LLM Recombined'), despite only fine-tuning the parameter equivalent of a single LLM.

\subsection{LEGO with Task-Dependent Data}
This experiment evaluates knowledge transfer in a task-dependent non-i.i.d. data distribution scenario. We use eight client SLMs with 75\% sparsity.
We split the databricks-dolly-15k dataset into each of its 8 domain areas, and each client model fine-tunes over one of these 8 segments. This means the client models do \textbf{not} have the same amount of local data, and each one only covers a \textbf{single} domain. Similar to the previous experiment, in each round a client will fine-tune on their local data and aggregate with the global LLM. A global update will be derived and applied to the client models.

The results of 8 task-dependent SLMs are shown in the last two rows of Table \ref{table:results}. The results highlight the advantage of LEGO for heterogenous data. LEGO outperforms FedIT as shown in the last column `LLM Recombined' with an accuracy of 0.571 versus 0.563.  Despite each SLM being fine-tuned on a different task, the knowledge transfers between models, resulting in a more robust global 0\% sparsity recombined LLM than any of the previous experiments. 

We additionally evaluate the effect of non-i.i.d. data on the quality of LEGO-produced global updates for SLMs. To do so, we track the performance of client SLMs over time, evaluating their average performance after every global update. 

\begin{figure}[h!]
    \centering
    \includegraphics[width=.8\linewidth]{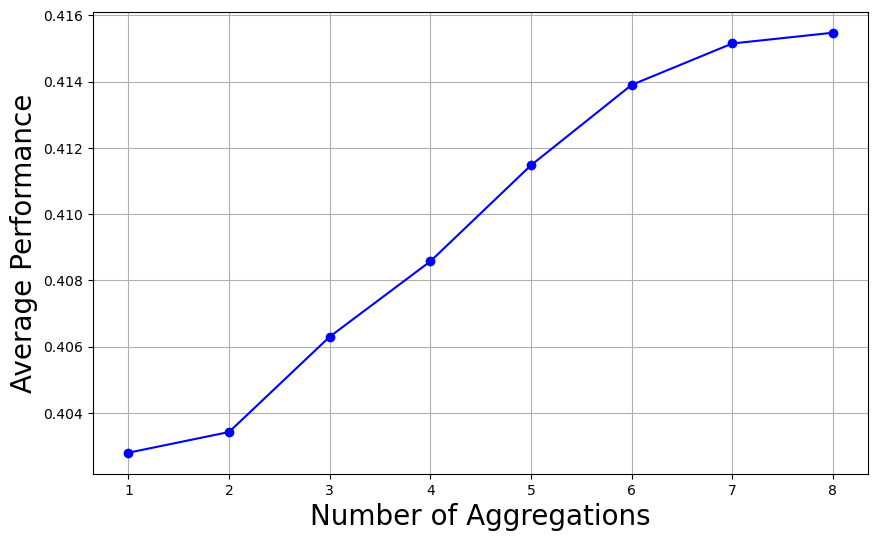}
    \caption{The performance of clients after each global update.}
    \label{fig:SLMlearning}
\end{figure}

Figure \ref{fig:SLMlearning} demonstrates that after every global update, the performance of the client SLMs increase almost linearly, despite the local data for each client not being the same length (with some \textbf{having $ 5\times$ the amount} of others). This demonstrates that SLMs are able to capture the knowledge from small amounts of data without underfitting, offsetting data heterogeneity.

\subsection{Combining Differently Shaped Blocks}

Just as not all (SLM) building blocks are the same size, they may not necessarily be the same shape. Regardless of the size or shape, the requirement is that they can stack together. LEGO demonstrates this principle.

\begin{figure}[h]
    \centering
    \includegraphics[width=.8\linewidth]{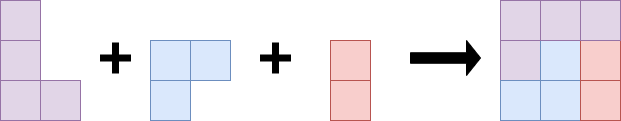}
    \caption{Combining differently shaped building blocks to create a larger block}
    \label{fig:exp3}
\end{figure}

Figure \ref{fig:exp3} shows three blocks of differing shapes being combined to create a new, larger block that encompasses the different shapes. The same can be done with SLMs, where each SLM can cover a different task or sparsity level, but be aggregated together to create a robust LLM that covers the diverse tasks of its components.

The results of the previous sections suggest that with LEGO, knowledge is transferred between SLMs fine-tuned with task-independent or task-dependent data. This section
investigates if this same knowledge transfer occurs between SLMs that are strictly of different shapes (sizes).

We initialize the FL environment with four heterogeneous client SLMs, each configured with a different sparsity level: \(0\%\), \(25\%\), \(50\%\), and \(75\%\). All SLMs, including the 0\% sparsity SLM (LLM), are fine-tuned on their respective local client data: one of four segments of i.i.d. data from the databricks-dolly-15k dataset. This segmentation is the same as Section \ref{iidexp}, ensuring that the local datasets are of equal size and cover the same domain areas in identical proportions. 

For this experiment, we first choose to fine-tune all the client models, and then we conduct a single round of aggregation, applying the resulting global updates uniformly across all client models.

\begin{table}[h]
\centering
\begin{tabular}{ccccc}
\hline
\textbf{Sparsity} & \textbf{0\%} & \textbf{25\%} & \textbf{50\%} & \textbf{75\%} \\ \hline
\textbf{Pruned}      & 0.559 & 0.554 & 0.529 & 0.384 \\
\textbf{Fine-Tuned}  & 0.563 & 0.561 & 0.526 & 0.412 \\ 
\textbf{Globally Updated}  & 0.568 & 0.565 & 0.542 & 0.396 \\ 
\hline
\end{tabular}
\caption{Average Model Accuracy over benchmarks for 4 SLM LEGO Components, each with different sparsity.}
\label{table:results2}
\end{table}

Table \ref{table:results2} displays the performance of the different-sized models. We benchmarked their performance at three different stages: when the LLM was initially pruned, resulting in the base SLM performance before fine-tuning (Pruned), when each client SLM is fine-tuned on its local data (Fine-Tuned), and the final adapters after the global updates (Globally Updated). As displayed in the table, we see that fine-tuning improves performance for all model sizes, with a significant performance gain at the 75\% sparsity level. 
The aggregation stage (Globally Updated row) improves performance for all models except the 75\% sparsity SLM.

The 75\% sparsity model's degraded performance after aggregation is likely due to the SLM's limited size. Previous work has shown that smaller models are better learners~\cite{turc2019well, raffel2020exploring}, creating an effect similar to dropout, forcing the limited neurons to create stronger and more general representations. During aggregation with the larger models, the small model's strongly learned representation becomes diluted by the larger model's weaker representation, degrading performance in the smaller model.

When comparing against the FedIT-produced baselines in Table \ref{table:results}, we see in Table \ref{table:results2} that the heterogeneous models produce an equally robust aggregated 0\% sparsity LLM, demonstrating successful knowledge transfer between models.

These results demonstrate that LEGO allows for flexible client model selection, enabling knowledge transfer between models of different sizes and tailoring client models to suit device capabilities, as opposed to being limited by the weakest client device.

To further understand the knowledge transfer between sizes, we repeat this experiment four times, each time omitting one of the client language models from the aggregation. This lets us view and analyze the individual contributions that each client model makes.

\begin{figure}[h]
    \centering
    \includegraphics[width=\linewidth]{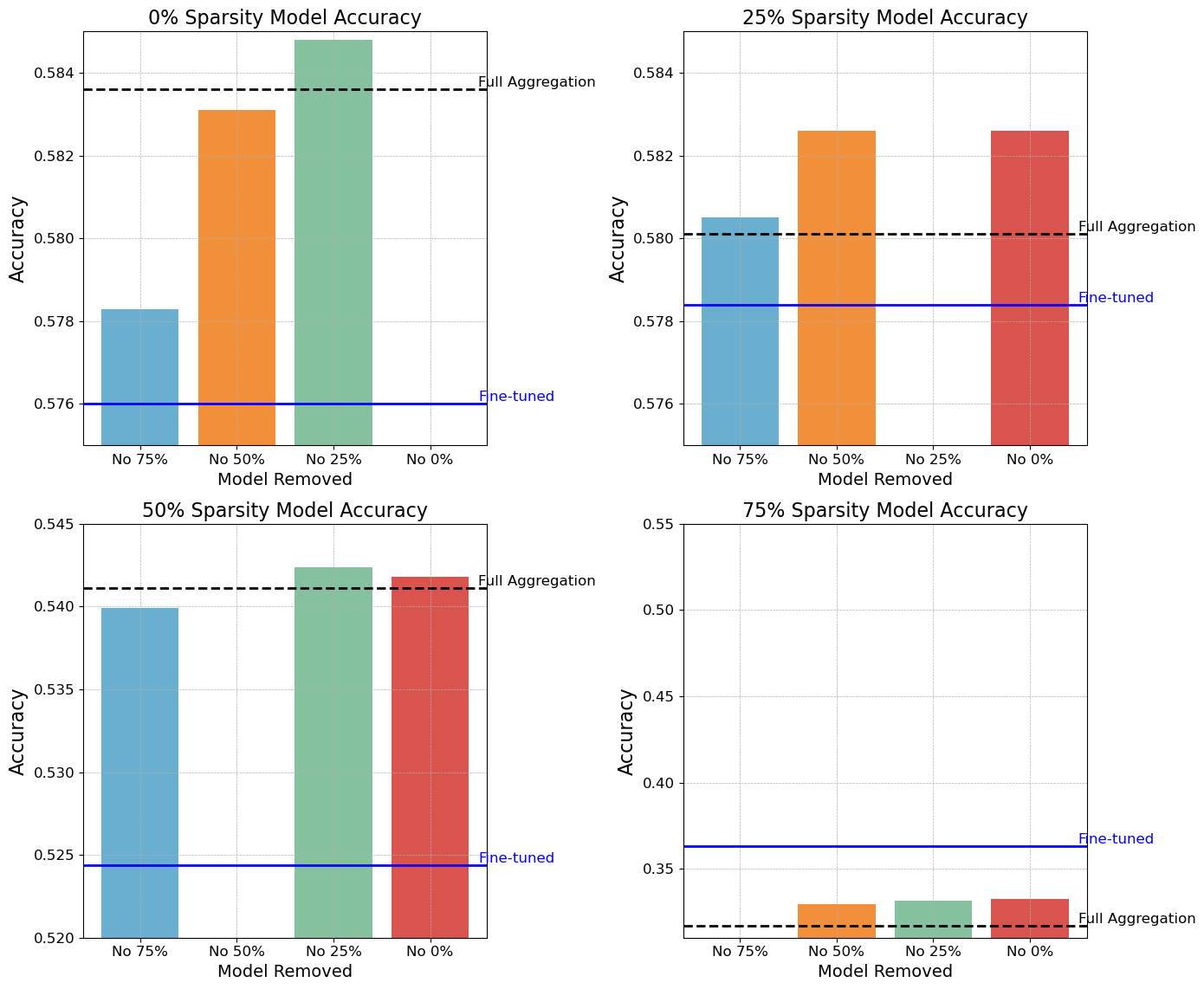}
    \caption{The accuracy of LEGO components on HellaSwag after aggregation with one omission. The solid blue line is the accuracy of the fine-tuned model, and the dotted black line is the globally updated performance, as listed in Table \ref{table:results2}.}
    \label{fig:abla}
\end{figure}

Figure \ref{fig:abla} shows that in the 0\% and 50\% sparsity models, performance degrades when the 75\% sparsity model is omitted from aggregation. These results demonstrate that LEGO allows for knowledge transfer from strictly smaller models to a larger model in an effective manner, confirming that smaller models are better learners. 

Despite any degradation relative to the globally updated performance from Table \ref{table:results2}, there is always an improvement over fine-tuning, except for the 75\% sparsity model. The 75\% sparsity model shows that as all aggregations degrade its performance, with the larger, denser models degrading it more. This confirms that larger models dilute its learned representation.

Given these results, we can come to two conclusions. First, smaller models create greater contributions to the 0\% sparsity LLM. Secondly, larger models do not transfer knowledge as effectively to smaller models.

While these experiments show potential for higher performance in a heterogeneous setting, the results are better underscored by already being on par with the FedIT baselines. This indicates that LEGO-produced models can further exceed the performance of their homogeneous counterparts.

\section{Related Work}\label{related}
Works on heterogeneous FL in the context of pretrained language models are sparse. 
We explain why this context is important in Appendix \ref{noCV}.
The first paper to cover model-agnostic FL in-depth was InclusiveFL~\cite{liu2022no}, where the authors used layer-pruned BERT models in a federated system and aggregated across layers. The authors found layer-pruning to have a negligible effect on BERT's performance - something that does not apply to modern LLMs due to large magnitude features distributed across the layers~\cite{dettmers2022gpt3}. We experimentally prove this in Appendix \ref{inclusivefl}.

We can extend this reasoning to similar approaches focused on layer selection that are only tested on encoder-style LLMs, like FedPepTAO~\cite{che2023federated}, since these all disregard large magnitude features.

We then look to homogeneous model FL applied to larger, decoder-style LLMs. FedIT~\cite{zhang2023towards} acts as the representation of traditional FL throughout our work, using FedAvg for aggregation as mentioned in Section \ref{experiments}. However, FedAvg cannot adapt to heterogeneous models, and as pointed out by other works, cannot account for heterogeneous ranks in the LoRA adapter\cite{bai2024federated}.

Newer works have continued to model themselves after FedIT's use of LoRA. Recently, enabling heterogeneous LoRA ranks in FL has been discussed in the literature. For example, FlexLORA computes a weighted average of LoRA adapters with different LoRA ranks, and then uses SVD for redistribution \cite{bai2024federated}. However, FlexLoRA assumes model homogeneity among client models, which is what allows for adaptive rank pruning in the LoRA adapter. 

The advantages of rank pruning \textbf{ do not} translate to the advantages of model pruning. Model pruning allows for more efficient fine-tuning and inference, whereas pruning LoRA only translates to more efficient fine-tuning, with the same inference costs as the initial LLM. Thus, in FlexLoRA, model selection is constrained by weakest device. In LEGO, pruning allows larger models (LLMs) to run on more powerful devices, and smaller models (SLMs) to run on weaker devices.

Additionally, their aggregation technique relies on multiplying each client's LoRA modules, $A$ and $B$, together, where $A \in \mathbb{R}^{r \times n}$ and $B \in \mathbb{R}^{n \times r}$. The multiplication results in the server creating the full-sized weights for every client model before aggregating them together. This extremely resource intensive operation limits the scalability of the technique relative to ours, where the LoRA modules stay separate. 

However, LEGO does not have to exclusively operate over PEFT adapters. The same approach and aggregation methods used for LoRA adapters can be performed with the actual client weights, or with the multiplied LoRA adapters. This means that rank-pruning techniques can be applied with or on top of LEGO, further decreasing SLM size, at the cost of increased computation for the server.

There are numerous other pruning techniques in the literature, but we choose to use the literature-standard pruning technique for our experiments, SparseGPT~\cite{frantar2023sparsegpt}. However, LEGO would be compatible with any pruning technique, including different client-structured pruning patterns, since LEGO preserves sparsity patterns in client models.

To the best of our knowledge, our work is the only work that looks to leverage pruned decoder-style LLMs for FL, allowing for fine-tuning \textbf{and} inference speedups for client models. Additionally, by pruning LLMs, we can scale client SLMs down to match heterogeneous client capabilities without limiting model size for computationally stronger clients. 

\section{Conclusions}\label{conclusions}
In this work, we have introduced LEGO, a building block methodology for federated fine-tuning of LLMs. By allowing for the use of pruned LLMs, we can use SLMs as task-specific learners for resource-constrained devices, and stack them into a fully robust LLM. This is enabled through our simple, yet effective, aggregation scheme, HeteAgg, which allows for the aggregation of heterogeneous SLMs. 
Through experimentation we demonstrate that LEGO can leverage SLMs, allowing for better adaptation to small amounts of data, stronger learning over non-i.i.d. client data distributions, and greater client flexibility by allowing for tailored client models for client devices. 
By enabling heterogeneous client resource budgets, LEGO creates a more scalable and resource-efficient FL system for private conversational AI.

\section{Limitations} \label{limits}
Our approach has limitations caused by prioritizing efficiency. As mentioned in Section \ref{method}, we operate over client LoRA adapters. Each LoRA module $A$ and $B$ is aggregated separately, which introduces noise to the resulting weights, as \[
 \underbrace{\sum A \times \sum B}_{\text{LEGO}} \;\;\; \neq  \underbrace{\sum (A \times B)}_{\text{Noise-Free Aggregation}.}
 \] Despite the noise, however, we show experimentally that LEGO produces robust models.

\bibliography{custom}
\bibliographystyle{acl_natbib}

\appendix

\section{Appendix}
\label{sec:appendix}

\subsection{Comparison of Pruning Methods}
\label{subsec:appenpruning}
As discussed in the Background section, there are two pruning techniques that dominate the literature. We test both SparseGPT and Wanda and analyze the best pruning technique to use. 

\setlength{\tabcolsep}{6pt}

\begin{table*}[h!]
\centering
\caption{Comparison of SparseGPT and Wanda Pruned Models}
\label{basepruning comparison}
\begin{tabular}{ccccc}
\hline
\textbf{Sparsity Level} & \multicolumn{2}{c}{\textbf{SparseGPT}} & \multicolumn{2}{c}{\textbf{Wanda}} \\ \cline{2-5} 
                        & \textbf{Pruned}      & \textbf{Fine-tuned}     & \textbf{Pruned}    & \textbf{Fine-tuned}   \\ \hline
0\%                     & 0.5694                & \textbf{0.5760}                 & 0.5694              & 0.5741               \\
25\%                    & 0.5654                & \textbf{0.5784}                 & 0.5672              & 0.5731               \\
50\%                    & 0.5144                & 0.5244                 & 0.5195              & \textbf{0.5377}               \\
75\%                    & 0.2989                & \textbf{0.3631}                 & 0.2692              & 0.2916               \\ \hline
\end{tabular}
\caption{All models were pruned from LLaMA-7B and evaluated over HellaSwag~\cite{zellers2019hellaswag}. The Fine-tuned models were fine-tuned over databricks-dolly-15k. Bolded scores are the best in sparsity level.}
\end{table*}

The results in table \ref{basepruning comparison} show that SparseGPT produces more robust models on average, with a significant advantage at higher levels of sparsity. However, SparseGPT is more computationally expensive when pruning, while Wanda is computationally inexpensive.

This provides us a few insights. 
The first is that regardless of pruning strategy, performance degrades significantly beyond 50\% sparsity. 
The second is that while more computationally expensive, SparseGPT may be necessary at high sparsity levels or more resource constrained client devices, as it not only produced a more robust model, but the increase in performance due to fine-tuning was almost double that of Wanda. 

Given these insights, the superior pruning method depends on the use case scenario. If we are defining rigid model sizes and assert that client devices will be initialized with one of these 'default' model sizes, then SparseGPT would be superior. This is especially true given our compute budget is capable of fine-tuning LLMs and performing inference, since SparseGPT is relatively cheap compared to those tasks if not being performed for ever device initialization. Thus, we can use SparseGPT to generate various model sizes/sparsity's before the FL process begins, and assign models accordingly.

However, in practice, creating a methodology to calculate the ideal model size given the device's compute budget would return more robust client models for users in the FL system. In this scenario, when a client is initialized, a model would be pruned according to their compute budget, meaning a lightweight process like Wanda would be superior.

However it is worth noting that, with the exception of high sparsity scenarios, the difference between the two pruning method's performances is negligible. Therefore, our results should be generalizable to both pruning methods.

Additionally, as pruning methods continue to evolve, the performance of pruned models will improve. Therefore its important evaluate model performance in our experiments with the limitations of current pruning techniques, but as pruning techniques improve, our methodologies and results would generalize to them and should scale accordingly.

In order to confirm if our experimental results are generalizable to other pruning techniques, we also test the Wanda-pruned SLMs for our HeteAgg experiment. We perform the same experiment involving 4 models at different sparsity levels, with its results displayed in table \ref{table:wanda4client}.

\begin{table*}[t]
\centering
\caption{Performance of Wanda pruned models on HellaSwag~\cite{zellers2019hellaswag}}
\label{table:wanda4client}
\begin{tabular}{cccc}
\hline
\textbf{Sparsity Level} & \textbf{Pruned} & \textbf{Fine-Tuned} & \textbf{Aggregated} \\ \hline
0\%                           & 0.5694           & 0.5741            & \textbf{0.5799}              \\
25\%                          & 0.5672           & 0.5731            & \textbf{0.5802}              \\
50\%                          & 0.5195           & 0.5377            & \textbf{0.5393}              \\
75\%                          & 0.2692           & \textbf{0.2916}            & 0.2717              \\ \hline
\end{tabular}
\end{table*}

\begin{figure}[h]
    \includegraphics[width=\linewidth]{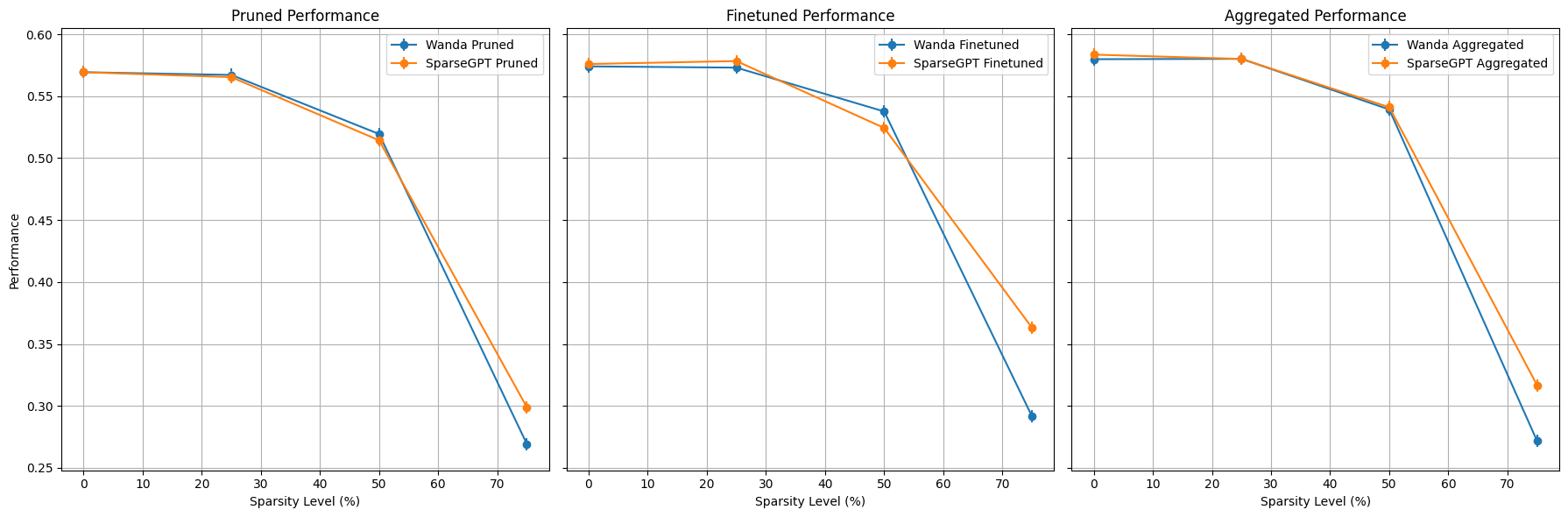}
    \caption{Performance of federated SparseGPT-pruned models relative to federated Wanda-pruned models when evaluated on HellaSwag~\cite{zellers2019hellaswag}}
    \label{fig:3graph}
\end{figure}

When plotted against SparseGPT's performance in figure \ref{fig:3graph}, we see that the effects of our FL approach are near identical. For sparsity $\geq50\%$, we see that the results are nearly identical, and the performance gap displayed by the fine-tuned 50\% sparsity SparseGPT-pruned model is corrected after model aggregation. 

While the performance on HellaSwag is different at high sparsity, that can be attributed to Wanda's weaker pruning ability at high sparsity levels. When viewing the Wanda and SparseGPT pruned 75\% sparsity models, we see the drop in performance due to aggregation after fine-tuning is nearly identical.  

Therefore, since the performance is nearly identical, and the only significant difference in performance can be attributed to the initial model performance as opposed to our FL system, we can generalize our FL method to other current pruning techniques.

\subsection{Experimental Comparison with InclusiveFL}\label{inclusivefl}

To confirm the effect of emergent large-magnitude features in LLMs discussed in Section \ref{related}, we experimentally compare InclusiveFL and layer pruning to LEGO and activation pruning. To do so, we layer-prune LLaMA-7B and modify our HeteAgg function to perform layer-wise aggregation.

We pruned LLaMA-7B to 24 and 16 layers, equivalent to 25\% and 75\% sparsity. We then put these two models and a 0\% sparsity LLaMA-7B model in the federated environment from Algorithm \ref{alg:fed_tuning_sparsity}, modifying the HeteAgg function to follow the pseudocode in the InclusiveFL paper. For the closest comparison we take select results from Section \ref{iidexp} and Table \ref{table:results}.

\begin{table*}[h]
\centering
\caption{Performance of layer-pruning~\cite{liu2022no} compared to activation pruning (our study).}
\label{table:llama-models-comparison}
\begin{tabular}{lcc|cc}
\hline
\multirow{2}{*}{\textbf{Sparsity / Layers}} & \multicolumn{2}{c|}{\textbf{Pruned}} & \multicolumn{2}{c}{\textbf{Fine-tuned \& Aggregated}} \\ \cline{2-5} 
                                        & \textbf{SparseGPT} & \textbf{Layer-Pruning} & \textbf{SparseGPT} & \textbf{Layer-Pruning} \\ \hline
Full Sized                              & 0.5694                  & 0.5694              & \textbf{0.5836}                  & 0.5148              \\
25\% Sparsity / 24 Layers                           & 0.5654                  & 0.3957              &\textbf{ 0.5801 }                 & 0.3658              \\
50\% Sparsity / 16 Layers                           & 0.5144                  & 0.3021              & \textbf{0.5411}                  & 0.3014              \\ \hline
\end{tabular}
\end{table*}

In Table \ref{table:llama-models-comparison}, we can see that even before federation, layer pruning fails to conserve model performance after pruning. This can be attributed to the emergent large-magnitude features in LLMs, as described in Section \ref{related} \cite{dettmers2022gpt3}. After federation, the fine-tuning and aggregation process degraded the performance, proving that this approach does not work for LLMs.

\subsection{Unique FL challenges}\label{noCV}

Federated Learning for LLMs comes with numeorus unqiue challenges. The overwhelming majority of model-heterogeneous FL literature is written in the context of computer vision (CV) based models. 

These previously proposed solutions cannot be applied to LLMs, due to challenges in language data and model architecture. 

The first major challenge is the difference in data. Language data and image data are fundamentally different. Text data is significantly more complex, being a high dimensional data where sequence context matter much more. In the context of transformer models, token dependencies across long sequences make many of the aggregation techniques discussed in CV much less effective.

Text data presents higher dimensional complexity than image data due to the sequential and contextual nature of language. Transformers must account for token dependencies across long sequences, making model compression and aggregation methods less effective than in vision models. This applies to ordered dropout from techniques like FjORD~\cite{horvath2021fjord} which relies on simple low dimensional input data for its DNNs. These techniques cannot capture the complex, dynamic token dependencies that transformers do.

Architecturally, language models and CNNs differ by more than just the use of transformers. CNNs, regardless of size, typically retain a similar architecture (convolutional and pooling layers). Most CNNs use most of their layers to learn how to "see", e.g. edge detection, color contrast, etc. This is why federated dropout or partial training methods work for CNNs. Methods like FedRolex, where stochasitc sub-model extraction occurs, rely on using CNNs because convolutional layers can function somewhat independently~\cite{alam2022fedrolex}.
Contemporary LLMs do not have such a standardized structure, where the layers can function independently.

This can be partially attributed to the large magnitude features in LLMs. These large magnitude features. These are parameters with significantly higher magnitudes than the rest of the parameters, and are sparse and distributed randomly across layers and have a significant effect on LLM performance. This is a phenomenon unique to transformers, something with which many FL methods do not adapt to.

This also extends to model compression. Pruning techniques such as filter and layer pruning are commonly used in CNNs, but are less effective for transformers, as demonstrated in appendix \ref{inclusivefl}. These techniques are often the basis of heterogenous FL techniques for CV or DNNs. Without pruning, many CNNs are small enough to communicate, like in

\subsection{Experimental Setup and Performance}\label{expsetup}

For all of the experiments, due to hardware limitations we use a client selection strategy that sequentially chooses clients. We use a client participation rate of 0.1, with a local batch size of 64 and only fine-tune for \textbf{one epoch}. For our LoRA adapter settings, we chose a rank and alpha of 16, and only target the q\_proj.

\begin{table*}[t]
\centering
\begin{tabular}{cccccc}
\hline
\textbf{Sparsity (\%)} & \textbf{Stage} & \textbf{HellaSwag} & \textbf{MMLU} & \textbf{SciQ} & \textbf{Arc} \\ \hline
\multicolumn{6}{c}{\textbf{4 SLMs With i.i.d. Data Distribution}} \\
\hline
0 & Pruned & 0.569 & 0.299 & 0.947 & 0.419 \\
0 & Aggregated & 0.586 & 0.294 & 0.944 & 0.447 \\
\\
75 & Pruned & 0.299 & 0.230 & 0.809 & 0.197 \\
75 & Aggregated & 0.364 & 0.294 & 0.944 & 0.4471 \\
\\
\hline
\multicolumn{6}{c}{\textbf{FedIT: 4 LLMs}} \\
\hline
0 & Aggregated & 0.575 & 0.286 & 0.956 & 0.453 \\
\\
\hline

\multicolumn{6}{c}{\textbf{8 Task-Dependent SLMs}} \\
\hline

0 & Pruned & 0.569 & 0.299 & 0.947 & 0.419 \\
0 & Aggregated &  0.586 & 0.298 & 0.953 & 0.446 \\
\\
75 & Pruned & 0.299 & 0.230 & 0.233 & 0.197 \\
75 & Aggregated & 0.359 & 0.241 & 0.813 & 0.233\\

\\
\hline
\multicolumn{6}{c}{\textbf{FedIT: 8 Task-Specific LLMs}} \\
\hline
0 & Aggregated & 0.570 & 0.279 & 0.951 & 0.452 \\
\\
\hline

\multicolumn{6}{c}{\textbf{4 Strictly Heterogeneous Models}} \\
\hline
0 & Pruned & 0.569 & 0.299 & 0.947 & 0.419 \\
0 & Fine-Tuned & 0.576 & 0.295 & 0.950 & 0.429 \\
0 & Aggregated & 0.584 & 0.301 & 0.953 & 0.435 \\
\\
25 & Pruned & 0.565 & 0.292 & 0.938 & 0.422 \\
25 & Fine-Tuned & 0.578 & 0.286 & 0.944 & 0.437 \\
25 & Aggregated & 0.580 & 0.295 & 0.944 & 0.442 \\
\\
50 & Pruned & 0.514 & 0.292 & 0.935 & 0.375 \\
50 & Fine-Tuned & 0.524 & 0.267 & 0.932 & 0.379 \\
50 & Aggregated & 0.541 & 0.292 & 0.932 & 0.404 \\
\\
75 & Pruned & 0.299 & 0.230 & 0.809 & 0.197 \\
75 & Fine-Tuned & 0.363 & 0.237 & 0.828 & 0.221 \\
75 & Aggregated & 0.317 & 0.229 & 0.832 & 0.206 \\
\\

\end{tabular}
\caption{Model Performance Across Different Configurations and Datasets}
\label{appenresults}
\end{table*}

Table \ref{table:results} showed the average model performance for each model. The individual results for each benchmark of each model is held in Table \ref{appenresults}. We evaluate each model on HellaSwag~\cite{zellers2019hellaswag}, MMLU \cite{hendrycks2021MMLU}, SciQ, and ARC~\cite{clark2018think}. We evaluate the models using the EleutherAI Language Model Evaluation Harness \cite{eval-harness}.

\end{document}